\documentclass[10pt,journal,compsoc]{IEEEtran}

\usepackage{graphicx}
\usepackage{tabularx}
\usepackage{caption}
\usepackage{subcaption}
\usepackage{multirow}
\usepackage{array}
\usepackage{mathtools}

\ifCLASSOPTIONcompsoc
  \usepackage[nocompress]{cite}
\else
  \usepackage{cite}
\fi

\begin{document}

\title{\title{Dynamic Facial Expression Recognition under Partial Occlusion with Optical Flow Reconstruction}}

\author{Delphine Poux, Benjamin Allaert, Nacim Ihaddadene, Ioan Marius Bilasco, Chaabane Djeraba and Mohammed Bennamoun
\IEEEcompsocitemizethanks{\IEEEcompsocthanksitem D. Poux (corresponding author), B. Allaert, I.M. Bilasco and C. Djeraba are with Centre de Recherche en Informatique Signal et Automatique de Lille, Univ. Lille, CNRS, Centrale Lille, UMR 9189 - CRIStAL -, F-59000 Lille, France.\protect\\
\IEEEcompsocthanksitem N. Ihaddadene was with Junia, Computer Science and Mathematics Department, F-59000 Lille, France.\protect\\
\IEEEcompsocthanksitem M. Bennamoun was with University of Western Australia (M002), 35 Stirling Highway, 6009 Perth, Australia.}
}

\markboth{}%
{Shell \MakeLowercase{\textit{et al.}}: Bare Demo of IEEEtran.cls for Computer Society Journals}

\IEEEtitleabstractindextext{%
\begin{abstract}
Video facial expression recognition is useful for many applications and received much interest lately. Although some solutions give really good results in a controlled environment (no occlusion), recognition in the presence of partial facial occlusion remains a challenging task. To handle occlusions, solutions based on the reconstruction of the occluded part of the face have been proposed. These solutions are mainly based on the texture or the geometry of the face. However, the similarity of the face movement between different persons doing the same expression seems to be a real asset for the reconstruction. In this paper we exploit this asset and propose a new solution based on an auto-encoder with skip connections to reconstruct the occluded part of the face in the optical flow domain. To the best of our knowledge, this is the first proposition to directly reconstruct the movement for facial expression recognition. We validated our approach in the controlled dataset  CK+ on which different occlusions were generated. Our experiments show that the proposed method reduce significantly the gap, in terms of recognition accuracy, between occluded and non-occluded situations. We also compare our approach with existing state-of-the-art solutions. In order to lay the basis of a reproducible and fair comparison in the future, we also propose a new experimental protocol that includes occlusion generation and reconstruction evaluation.
\end{abstract}

}

\maketitle

\IEEEdisplaynontitleabstractindextext

\IEEEpeerreviewmaketitle

\IEEEraisesectionheading{\section{Introduction}\label{sec:introduction}}

Facial expression recognition has many applications in several domains such as healthcare, security, sentiment analysis, and marketing. Specific applications include facial expression recognition-based sentiment analysis to monitor students' learning during e-learning sessions, or the automatic monitoring of the driver's temper to avoid car accidents. In order to bring life to these applications, several solutions have been proposed to automatically recognize expressions. 

These solutions give really interesting results in a constrained environment. However, in real-world situations, face occlusions caused by the hand of the person or with accessories (such as a surgical mask, a scarf, a hat, glasses, ...) occur frequently which make the recognition task more challenging.

In order to handle occlusions, many solutions have been proposed in the literature. Two categories can be roughly distinguished : methods that exploit the remaining visible information and methods that reconstruct the hidden parts of the face. The great advantage of the reconstruction is the fact that we get back to an ideal situation with the entire face from which the facial expression can be recognized. These methods are mainly based on texture or geometry.
However, motion has proven its effectiveness for facial expression recognition \cite{bassili1979},\cite{li2020}. In spite of these results, most of the solutions to handle occlusions are based on static images. Recent papers tend, in general, to neglect research on motion information as seen in the recent CVPR 2019 statistics\footnote{https://cvpr2019.thecvf.com/files/CVPR 2019 - Welcome Slides Final.pdf}. Nevertheless, motion remains an important and really useful information, as demonstrated for example, by the ECCV 2020 best paper award\footnote{https://eccv2020.eu/awards/} which proposed a new deep learning architecture to calculate optical flow \cite{teed2020raft}.

\begin{figure}[h!]
\centering
\includegraphics[width=\columnwidth]{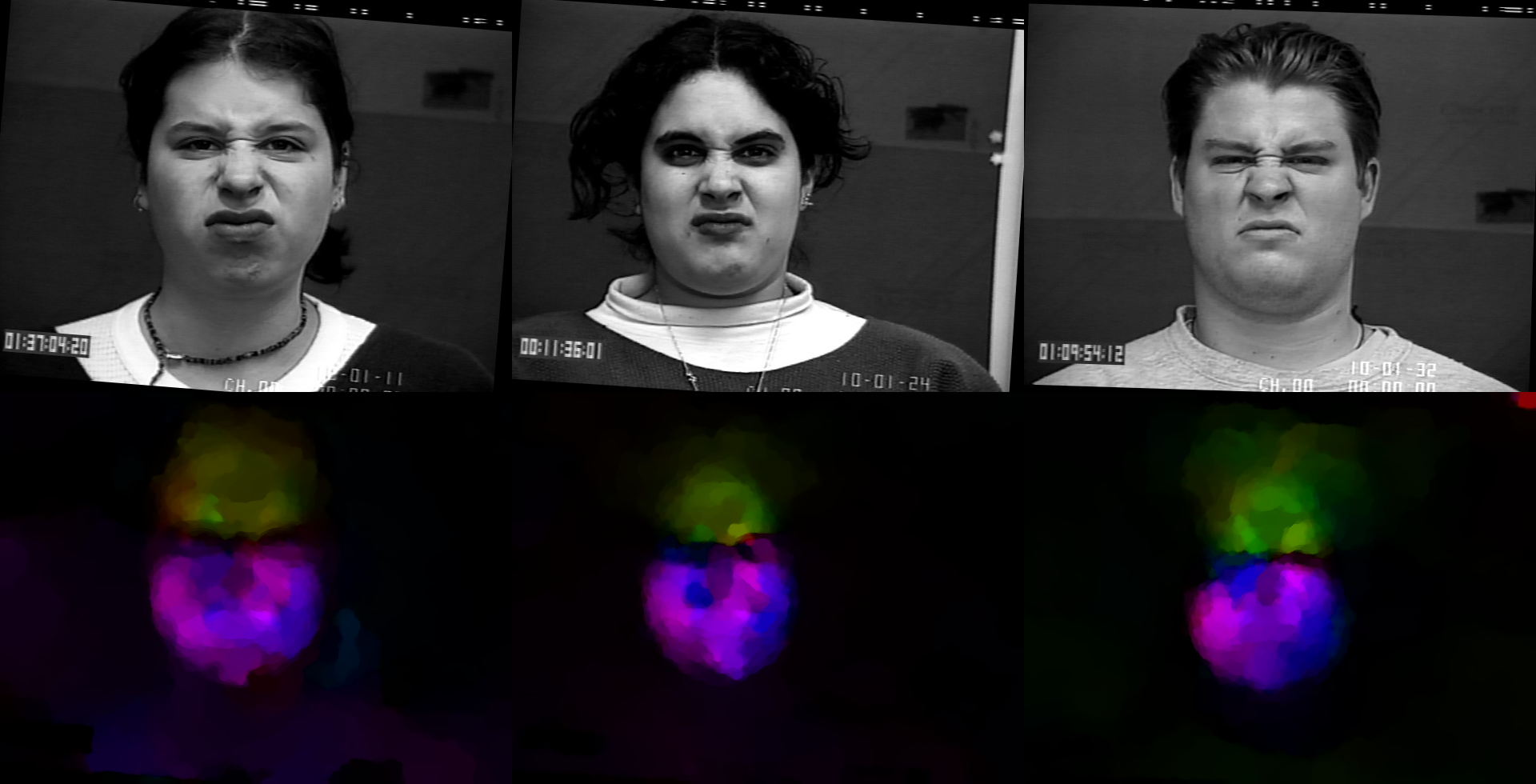}
\caption{
The first line shows the texture of several persons from the CK+ dataset having an expression of disgust and the second line shows the corresponding optical flows calculated between the first frame of the sequence and the last frame (shown in the first line here). 
This figure illustrates the great variability in terms of texture of several persons having the same expression on one hand, and, on the other hand, the similarity of the corresponding optical flows. To deal with occlusions, some solutions in the literature try to reconstruct the texture. In our work, we exploit the movement similarity between several persons, based on optical flows, and propose the reconstruction of the optical flows in the occluded regions.
}
\label{movement_similarity}
\end{figure}

In order to overcome these issues, we propose in this paper a solution that reconstructs the face movement instead of its texture or geometry. Our idea is based on the fact that, due to the propagation of the face movement, the motion information that remains available in the visible face part can be very useful for the reconstruction of the occluded parts.

Moreover, there is a strong similarity of the face movement for the same expression produced by different persons, as illustrated in Fig. \ref{movement_similarity}. This observation highly enhances the reconstruction of the lost information. This observation of similarity also tends to show that movement is more suitable because it is not dependent on the identity of the person which implies a reduction of the motion variability between different persons doing the same expression.

\begin{figure}[!h]
    \includegraphics[width=\columnwidth]{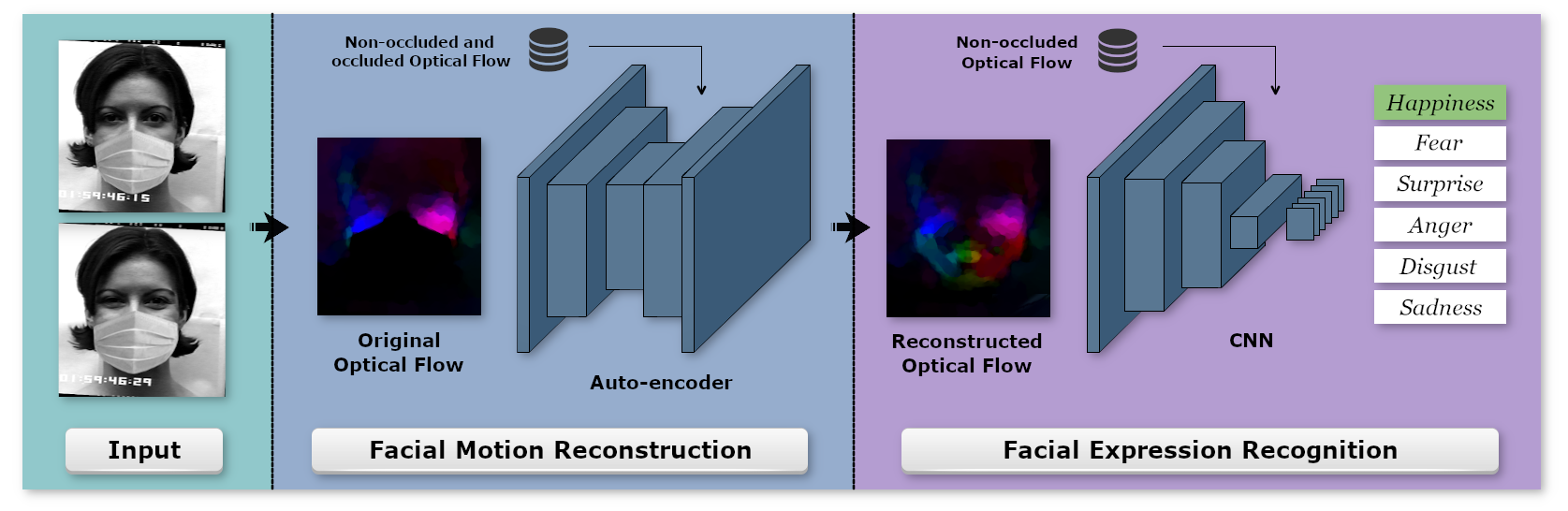}
    \caption{The entire process of our approach consists in three main steps. The first step consists in calculating optical flows between the frames of a sequence of occluded face. The second step consists in reconstructing optical flows corrupted by the occlusion using a trained auto-encoder. The auto-encoder is trained using pairs of occluded and non-occluded optical flows. The restored optical flows are then used directly for the classification step to predict the expression.}
    \label{fig:entireProcess}
\end{figure}

The proposed method is based on an auto-encoder architecture which has proven its effectiveness on similar applications. Our solution, illustrated in Fig. \ref{fig:entireProcess}, is a two-stage process. The reconstruction module, takes as input an optical flow calculated from two frames of an occluded face and restore, using an AE, the optical flow of the occluded parts. The recognition module takes then the restored optical flow as input in order to predict the expression class.

To evaluate the proposed solution, the difficulty encountered is the variability of the evaluation protocols used in the literature. Indeed, the methods in the state-of-the-art are validated using different datasets, different data splitting protocol, and different occlusions (in terms of position, size and texture). Thus, we propose here a clear and available evaluation protocol to make possible a fair comparison with our method and other methods in the future. To reproduce the proposed protocol, the code for occlusion generation is available on demand.

So, our proposal brings two novelties: 

\begin{enumerate}
    \item To the best of our knowledge, this is the first work that directly reconstructs the movement (optical flow) for facial expression recognition. 
    \item We propose a new and reproducible experimental protocol  with a fixed split of dataset for training and testing, and a clear and precise occlusion generation process.
\end{enumerate}

The next section gives a review of existing methods in the literature that handle occlusions. An overview of the different experimental protocols (datasets, occlusions, etc) used by researchers is then presented in Section \ref{scope}. Our method is then explained and detailed in Section \ref{approach}. The evaluation protocol and the results are presented in Section \ref{evaluation}. And, we prove that movement reconstruction improve expression reconstruction in presence of occlusions. Then, we conclude in Section \ref{subsec:conclusion}.

\section{Related works}
\label{scope}
The state-of-the-art will be addressed according three notions: the methods proposed to handle occlusions, the experimental protocol used for the evaluations and the facial expression recognition methods. 

\subsection{Methods for Facial Expression under Occlusion}
\label{subsec:methods}
In the literature, several methods have been proposed to handle occlusions. These methods can be classified in two classes: the methods that exploit the remaining visible regions of the face and the methods that reconstruct the occluded parts. 

\textit{Methods that exploit the remaining visible regions of the face.} The attention is focused on the sub-regions that belong to only the visible part of the face. Liu et al. \cite{liu2014b} proposed a method that divides the face in a grid of six non-overlapping regions. A classification process is used for each region based on Weber Local Descriptor (WLD) features and an argmax decision fusion is applied. Nevertheless, a grid cutting restricts the algorithm robustness to some specific type of occlusions. Moreover, by using the decision fusion strategy, the authors assume that the regions are completely independent from each others i.e., no correlation between the different facial regions to distinguish the different facial expressions. In 2018, Dapogny et al. \cite{dapogny2018confidence} proposed a method that calculates features from different regions extracted from the face. The authors also proposed an auto-encoder trained to determine a confidence weight, which corresponds to the occlusion rate, of each region. A weighting scheme is then used to weight the different region features before classification. The regions are first classified independently then a fusion decision is performed. These different solutions are based on still images and do not exploit temporal aspects. Poux et al. \cite{poux2020facial} have hypothesized that movement could bring more useful information and, especially, dense movement on which a property of propagation can be useful in the case of occlusion. To exploit this property, they propose to build a facial framework specific to the different partial occlusions from an initial facial framework of twenty-five regions. The calculated facial frameworks are subsets of regions for each facial expression optimal to recognize facial expressions in spite of the occlusion. These methods are based on the manual division of the face more or less adapted to facial expressions. These divisions are thus not flexible enough to handle a wide range of occlusions. To exploit visible regions more accurately according to occlusions, some methods based on attention mechanism are proposed \cite{li2018,wang2020}. These methods focus the attention on the most informative parts of the face and the features are weighted according to the results of the attention map. This mechanism is particularly adapted for the case of partial occlusions as it automatically detects the non informative parts which here corresponds to occluded parts.

\begin{figure}[!h]
    \centering
    \begin{subfigure}[b]{0.2\textwidth}
        \includegraphics[width=\textwidth]{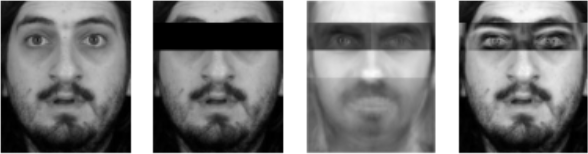}
    \end{subfigure}
    \hspace{2em}
    \begin{subfigure}[b]{0.2\textwidth}
        \includegraphics[width=\textwidth]{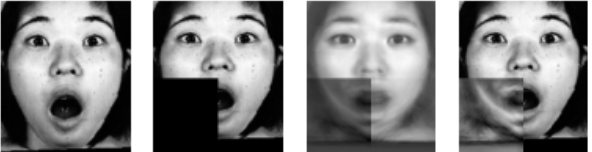}
    \end{subfigure}
    \caption{Results of reconstruction of the face extracted from \cite{cornejo2018emotion} proposed by Cornejo et al. In the first column of each example, images are extracted from MUG and JAFFE datasets. In the second column, the images have been occluded. In the third column, we have the resulting reconstruction from the RPCA algorithm. Finally, the last column is obtained by filling in the occluded parts of the face by the reconstructed one.}
\label{artefactsRPCA}
\end{figure}

\textit{Methods that reconstruct the occluded parts.} With the recent progress in neural networks architectures, the proposed solutions have known different improvements. Some first solutions have been proposed to reconstruct features according to the visible ones \cite{towner2007,zhang2015adaptive}. These first solutions mostly reconstruct facial landmarks lost by the occlusion. This kind of solutions are highly dependent on facial landmarks detection. However, landmarks detection in presence of facial occlusion remains today a challenging task \cite{wu2019facial}. To get around this dependency, solutions propose to reconstruct directly the texture of the face. The PCA algorithm, initially proposed to reduce dimension is frequently used to reconstruct faces. In the case of occlusions, RPCA has proven to be more robust to occlusions and allows a reconstruction of an unoccluded face. Some solutions are proposed with an RPCA reconstruction as a first step in order to get an unoccluded face \cite{cornejo2018emotion}. While it is an often used solution, RPCA tends to add some artefacts on reconstructed images as seen in Fig. \ref{artefactsRPCA}, these artefacts may have an impact on the facial expressions of the persons and on the classification process.

Some solutions have been proposed based on neural network architectures which is more flexible to add some constraints in the reconstruction. A first architecture have been proposed in 2011 by Ranzato et al. \cite{ranzato2011deep} to reconstruct occluded faces thanks to an extension of a DBN algorithm. Recent advances in neural network architectures have proven their effectiveness while reducing handcrafted processes. Among these architectures, the auto-encoder, which, as the PCA, was initially proposed to reduce the dimensionality and calculate some features from an unsupervised way, turned out to be also particularly adapted to reconstruct corrupted images by adding a decoder. Indeed, the training step of this kind of architectures consists in supervising the reconstruction in order to find the most discriminative features in the bottleneck of the architecture. Based on this kind of architectures, GAN architectures have been proposed in order to get a more realistic output image. A GAN architecture is composed of two adversarial networks : a generator which generates new data and a discriminator initially trained to recognize false and real images. This kind of architectures are now used in solutions to reconstruct occluded faces \cite{li2017generative} in use cases such as facial recognition but are still poorly used for facial expression recognition. Recently, Lu et al. \cite{lu2019wgan} proposed a WGAN architecture used to reconstruct partially occluded faces composed of a generator based on an auto-encoder architecture and two discriminators : one classical discriminator trained to distinguish real or face images and another trained to recognize facial expressions. While these architectures have proven their effectiveness, GANs are complex architectures to set-up, require a huge amount of data to be trained and the presence of two networks makes the stability of the entire architecture highly complex to find.

This complexity can be explained by the fact that with this kind of architectures, we are searching for a realistic image, which implies a coherent texture. This coherence is partly linked to the identity of the person and is complex to find. Moreover, in the case of facial expression recognition, besides this coherence of texture, the architecture has to reconstruct the appropriate facial expression.

In summary, two categories of methods are proposed in the literature to recognize facial expressions in the presence of occlusions. In these two categories, dense movement is poorly used even though it seems particularly adapted. The similarity of the movement between different persons tends to prove the independence of the movement from the identity of the person. This property seems to bring a real asset to reduce the complexity of GANs architectures by abstracting from texture coherence in order to focus on an information highly correlated with facial expressions. As the coherence of the texture is not as important in the field of movement features, the adversarial network seem to be less adapted and the auto-encoder seems to be sufficient.

\subsection{Experimental Evaluations}
\label{subsec:evaluations}
Evaluation protocols proposed in the literature are various and it is, for the moment, difficult to reproduce experimental protocols and compare results with state-of-the-art methods. Indeed, there is a variability in terms of datasets, occlusions and protocols.

\subsubsection{Datasets}
In this first part, we propose to compare the datasets used in experimental protocols of the solutions exposed in the previous paragraph.

Table \ref{tab:datasets_in_soa} summarizes different datasets used to evaluate methods in the literature. As seen in this table, the datasets are various. This variety is a first barrier to compare the effectiveness of the different methods proposed in the literature and to identify a common validation protocol.

\begin{table}[h!]
    \begin{tabularx}{3.5in}{|l|X|}
    \hline
    \textbf{Paper} & \textbf{Datasets}  \\
    \hline
    Towner and Slater \cite{towner2007}, 2007 (2) & CK \\
    \hline
    Ranzato et al. \cite{ranzato2011deep}, 2011 (2) & CK, TFD \\
    \hline
    Liu et al. \cite{liu2014b}, 2014 (1) & JAFFE \\
    \hline
    Zhang et al. \cite{zhang2015adaptive}, 2015 (2) & CK, JAFFE \\
    \hline
    Dapogny et al. \cite{dapogny2018confidence}, 2018 (1) & CK+, BU-4D, SFEW\\
    \hline
    Cornejo et al. \cite{cornejo2018emotion}, 2018 (2) & CK+, JAFFE, MUG \\
    \hline
    Li et al. \cite{li2018},2018 (1) & RAF-DB, AffectNet, CK+, MMI, Oulu-Casia, SFEW \\
    \hline
    Lu et al. \cite{lu2019wgan}, 2019 (2) & AffectNet, RAF-DB \\
    \hline
    Poux et al. \cite{poux2020facial}, 2020 (1) & CK+ \\
    \hline
    Wang et al. \cite{wang2020}, 2020 (1) & FERPlus, AffectNet, RAF-DB, SFEW \\
    \hline
\end{tabularx}
\caption{Datasets used to evaluate different methods to handle occlusions in the literature and the associated category of method. The categories correspond to : (1) : Exploitation of the remaining visible regions of the face for facial expression recognition, (2) : Reconstruction of the occluded parts of the face for facial expression recognition.}
\label{tab:datasets_in_soa}
\end{table}

Table \ref{tab:datasets_apparitions} gives some additional information about the mostly used datasets in the literature. In this table, we only keep the datasets which appear at least twice in Table \ref{tab:datasets_in_soa}. As CK+ is an extension of the CK dataset, we have decided to count only one dataset. In this table, we also report the type of data : videos or still images. We can remark that, because we are working on movement features, in addition to be the most representative dataset used for evaluation in the literature, CK+ is also the most appropriate to study movements. In this table, we can notice that methods are evaluated both on controlled and uncontrolled datasets. On controlled datasets, the subjects keep their faces in front of the camera, there is no head pose variation, a controlled illumination, no occlusion and the expressions are acted. Nevertheless, we can also notice that evaluations on controlled datasets are slightly more frequent than on uncontrolled datasets. Controlled environment is, in fact, more appropriate to focus on the challenge of occlusions. It, in fact, allows to have the entire control on the occlusions studied and to ensure the results obtained are due to the occlusion and not the data itself. To know the impact of the occlusions it is then possible to compare the results with those obtained on unoccluded data.

\begin{table}[h!]
    \centering
    \begin{tabular}{|c|c|c|c|}
    \hline
    \textbf{Dataset} & \textbf{Apparitions} & \textbf{Video sequences} &    \textbf{Controlled}  \\
    \hline
    CK-CK+ & \textbf{7} & \textbf{yes} & \textbf{yes} \\
    \hline
    JAFFE & 3 & no & \textbf{yes} \\
    \hline
    AffectNet & 3 & no & no \\
    \hline
    SFEW & 3 & no & no \\
    \hline
    RAF-DB & 3 & no & no \\
    \hline
    \end{tabular}
    \caption{Comparison of the mostly used datasets in the literature.}
    \label{tab:datasets_apparitions}
\end{table}

We propose in the next section to review the simulated occlusions in the literature. Because CK+ is the mostly used in the literature and the most appropriate to study movements, we focus the analysis on papers working on this dataset.

\subsubsection{Occlusion Generation}
While the datasets used to evaluate methods are various, the CK+ is slightly more represented than others. As CK+ is a completely controlled dataset, occlusions have to be simulated. In this section, we compare different simulations proposed in the literature. Table \ref{tab:occlusions_sim} exposes different simulations applied on CK+. As seen in this table, simulations in the literature are various in term of localisation and simulation of these occlusions. We can nevertheless notice that, in spite of different sizes and localisations, occlusions of the eyes and the mouth regions seem to be frequently applied.

\begin{table}[h!]
\centering
\begin{tabular}{| >{\centering\arraybackslash} m{1.8cm} | >{\centering\arraybackslash} m{3cm} | >{\centering\arraybackslash} m{3cm}|}
\hline
\textbf{Paper} & \textbf{Occlusion localisation} & \textbf{Occlusion simulation} \\
\hline
Dapogny et al. \cite{dapogny2018confidence} & \centering \includegraphics[width=0.6cm]{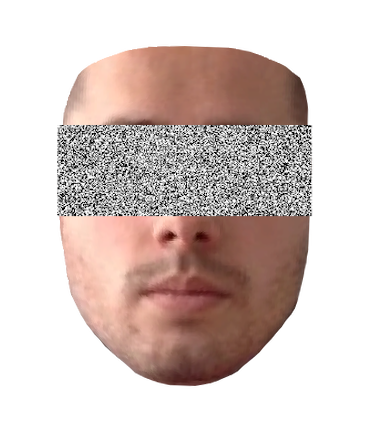} \includegraphics[width=0.6cm]{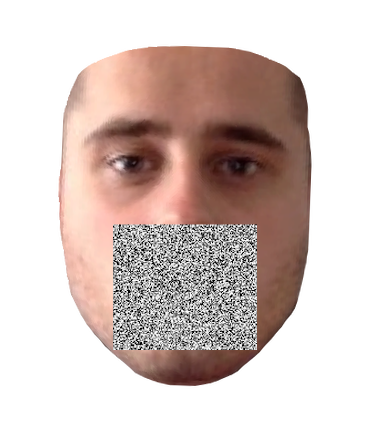} & Overlay of a noisy pattern\\
\hline
Poux et al. \cite{poux2020facial} & \centering \includegraphics[width=0.6cm]{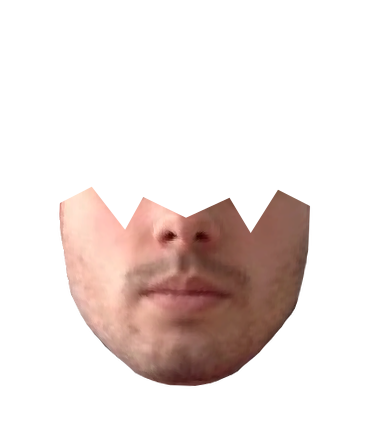} \includegraphics[width=0.6cm]{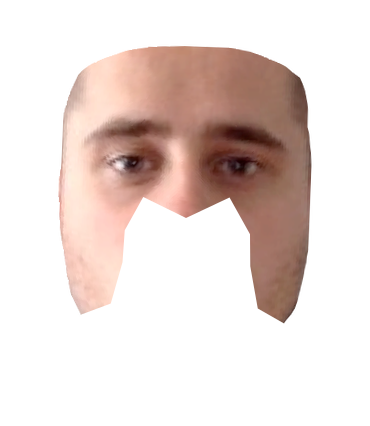} \includegraphics[width=0.6cm]{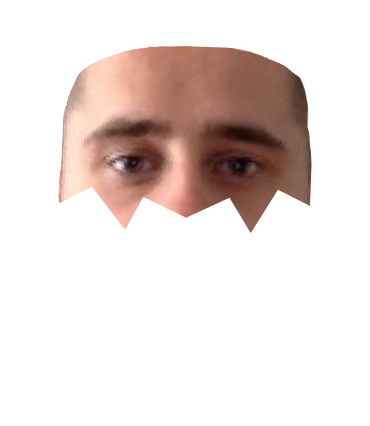}  & Removal of features \\
\hline
Li et al. \cite{li2018} & \centering \includegraphics[width=0.6cm]{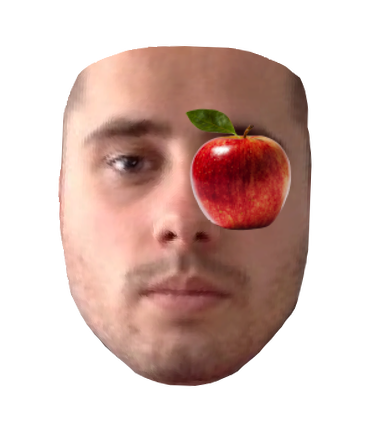} & Overlay of images of objects \\
\hline
Cornejo et al. \cite{cornejo2018emotion} & \includegraphics[width=0.6cm]{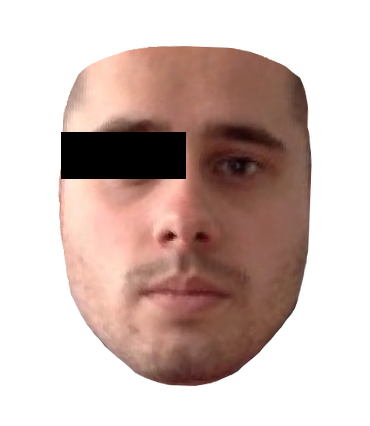} \includegraphics[width=0.6cm]{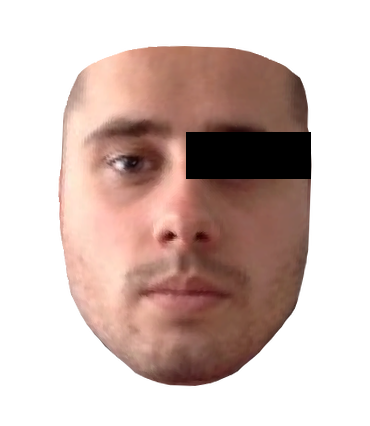} \includegraphics[width=0.6cm]{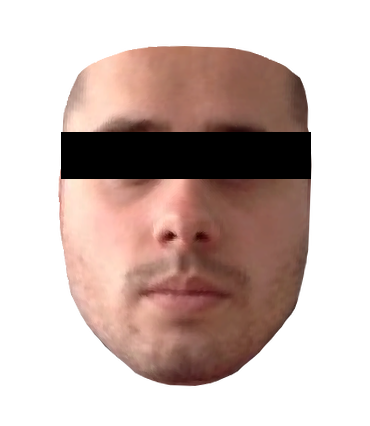} \includegraphics[width=0.6cm]{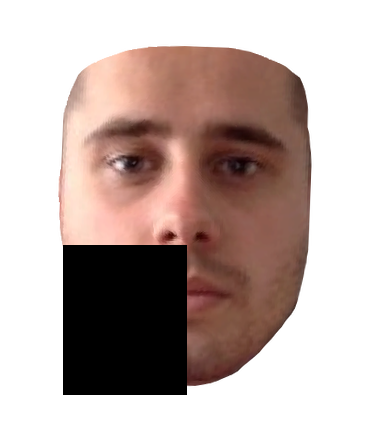} \includegraphics[width=0.6cm]{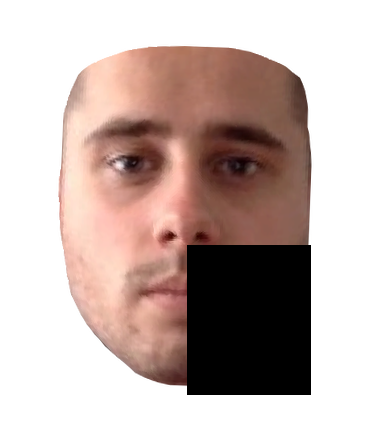} \includegraphics[width=0.6cm]{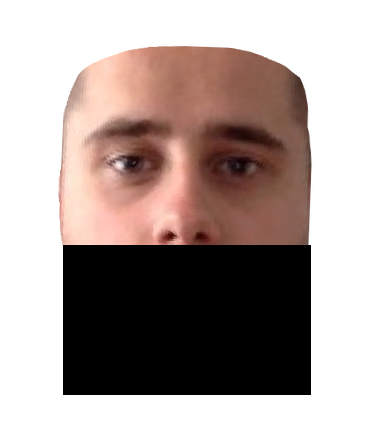} & Overlay of black boxes\\
\hline
\end{tabular}
\caption{Summary of different occlusion simulations on the CK+ dataset.}
\label{tab:occlusions_sim}
\end{table}

To get an entire experimental protocol, we propose next to compare the evaluation protocol used to evaluate these methods.

\subsubsection{Experimental Protocol}
The previous sections show that a comparison of methods is complex because of a variety in terms of datasets used and occlusion simulations. In these sections, we have seen that the CK+ dataset seems to be majoritarily used and, in spite of the variety of occlusions, areas of the eyes and on the mouth regions are mainly occluded. Finally, we propose in this last part to compare the experimental protocols used in the different papers evaluated on the CK+ dataset. Table \ref{tab:protocols} shows that, except for Cornejo et al. \cite{cornejo2018emotion}, the 10 folds cross validation protocol seems to be a representative protocol used in the literature.

\begin{table}[h!]
    \centering
    \begin{tabularx}{3.5in}{|c|X|}
    \hline
        \textbf{Paper} & \textbf{Experimental protocol} \\
        \hline
        Dapogny et al. \cite{dapogny2018confidence} & Out of bag error (OOB) \\
        \hline
        Poux et al. \cite{poux2020facial} & 10 folds cross validation \\
        \hline
        Li et al. \cite{li2018} & 10 folds cross validation \\
        \hline
        Cornejo et al. \cite{cornejo2018emotion} & 80\% for training, 20\% for testing. 50\% of training data is occluded, 50\% of testing data is occluded \\
        \hline
    \end{tabularx}
    \caption{Protocols applied to evaluate different methods on the CK+ dataset.}
    \label{tab:protocols}
\end{table}

Evaluations in the literature do not allow a fair and easy comparison. Indeed, evaluations vary in terms of datasets, occlusion simulations and protocols. However, we can notice that the CK+ dataset seems to be mostly used in the literature. This dataset is, in fact, an appropriate dataset as it is a controlled dataset with no particular challenge and, thus, allows to focus on the challenge of occlusions. Moreover, simulating occlusions allows a comparison between the results obtained with and without occlusions as we have ground truth images without occlusion. If we focus on the simulation of the occlusions, we notice that the occlusions are various in terms of localisation and simulation even if, generally, mouth and eyes occlusions are frequently studied. Finally, the evaluation of the CK+ dataset is frequently based on a 10-folds cross validation protocol.

In this paper, we propose to exploit the performance of auto-encoders to reconstruct directly the optical flow calculated from occluded faces for facial expression recognition by taking advantage of the similarity of the movement of the several expressions between different persons. To evaluate our solution, we propose a new reproducible experimental protocol with the CK+ dataset, available occlusions of the eyes and mouth regions and a 10-folds cross validation protocol.

\section{Proposed approach}
\label{approach}
In this paper, we propose to take advantage of the performances of generative algorithms to reconstruct occluded facial expressions by reconstructing directly the occluded optical flow. In order to do that, we propose a denoising auto-encoder which, takes the optical flow calculated between two occluded images as input and returns the unoccluded flow as output. We underline the fact that, even if generative algorithms have been used in the literature to reconstruct face images, to the best of our knowledge, our approach is the first to deal with the movements of the face to perform facial expression recognition in the presence of occlusions. 

\subsection{Data preparation}
For our data preparation, we consider the publicly available datasets for facial expression recognition, and the facial occlusion approaches that are available in the literature, to propose a more suitable testing protocol. First, we combine publicly available datasets which deal with expression recognition in the presence of occlusions. The advantage combining several datasets offers the ability to train our proposed approach on larger datasets. Second, we apply occlusions on several parts of the face. Finally, we calculate normalized optical flows.

\subsubsection{Occlusions Generation}

\begin{figure}[h!]
\centering
\includegraphics[width=2.5in]{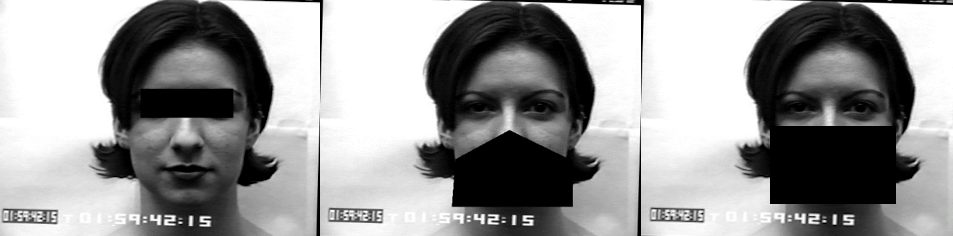}
\caption{An example of generated occlusions, applied on an image from the CK+ dataset, and used in our evaluation.}
\label{occlusions}
\end{figure}

In order to cope with most of the occlusions already studied in the literature \cite{zhang2018}, we simulate occlusions around the eyes and the mouth, as illustrated in Fig. \ref{occlusions}. Eyes and mouths are important parts for facial expression recognition \cite{kotsia2008analysis}.

\subsubsection{Optical Flow Calculation}

\begin{figure}[h!]
\includegraphics[width=3.5in]{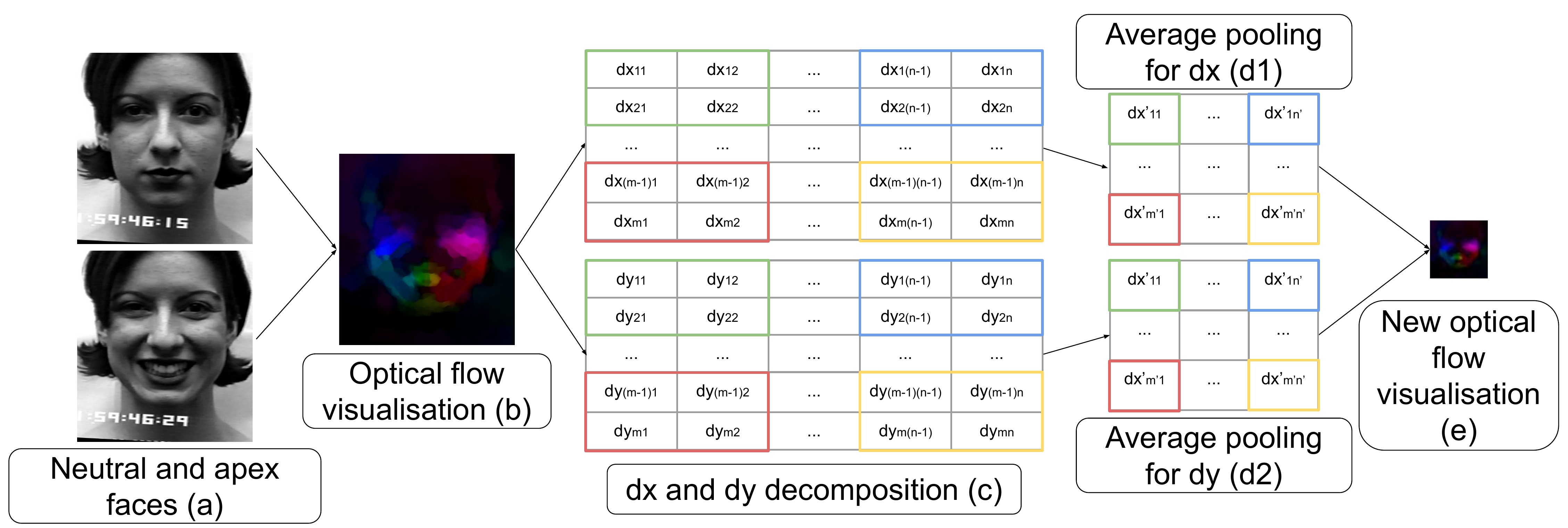}
\caption{This figure illustrates the calculation process of the normalized optical flow. The two images of Fig. \ref{optflow_resize}-(a) are the first and the last of a video sequence of the CK+ dataset where the original images have been cropped automatically with respect to the eyes positions and the inter pupilar distance to only obtain the inner face region. Fig. \ref{optflow_resize}-(b) is the visualisation in the HSV space of the optical flow calculated between the two images of Fig. \ref{optflow_resize}-(a) and Fig. \ref{optflow_resize}-(c) the matrix representation of the $dx$ and $dy$ decomposition which correspond respectively to the horizontal and vertical displacements of pixels. An average pooling is applied on these two matrices as represented in $d_1$ and $d_2$. In the example, the initial matrices are $mxn$ matrices reduced to $m'xn'$ matrices. Each $dx'$ and $dy'$ corresponds to the mean of the values in the corresponding region. A visualisation of this reduction is shown with Fig. \ref{optflow_resize}-(e) which corresponds to a reduction of the optical flow visualised in Fig. \ref{optflow_resize}-(b).}
\label{optflow_resize}
\end{figure}

In order to keep as much information as possible during the calculation of the optical flow, we calculate the flow on the original images (i.e., we keep their initial resolution) and we directly scale the optical flows in order to satisfy the requirement of normalized input for the deep networks that are used in our recognition and reconstruction steps. For that purpose, we propose the three steps process illustrated in Fig. \ref{optflow_resize}. First, we process the images in their original sizes. Second, we crop the face 
based on the positions of the eyes and the inter pupilar distance. Third, we compute the optical flow from the cropped faces using the Farnebäck \cite{farneback2003} method. We have selected the Farnebäck method to compute optical flows as Allaert et al. \cite{allaert2019} show that this method is particularly adapted for facial expressions. In order to generate normalized optical flows for all images and to reduce the computational cost of the auto-encoder, we resize the optical flows to a smaller size in the $x$ and $y$ dimensions. The evaluation of the optimal parameter size is covered in the evaluation section (Section \ref{evaluation}). Each new value of $x$ and $y$ is calculated based on the elements of a sliding window in the original flow. For each coordinate $(i,j)$, the new value is calculated using the following equation : 

\begin{equation}
\begin{split}
resize & (OF,(i,j)) = \\
 & (\mu OF[(\lfloor dx*i \rfloor,...,\lfloor dx*(i+1) \rfloor \lceil *dx \rceil - 1)],\\
 & \; \mu OF[(\lfloor dy*i \rfloor,...,\lfloor dy*(i+1) \rfloor \lceil *dy \rceil - 1)]) \\
\end{split}
\end{equation}

\noindent where $dx$ and $dy$ are the coefficients between the original and the final size ($dx=origSize\_x / finalSize\_x$ and $dy=origSize\_y / finalSize\_y$) and $\mu$ represents the average of the optical flow values in the window. $origSize\_x$ and $finalSize\_x$ are the original and final widths of the image, respectively.

\subsection{Optical Flow Reconstruction}

\subsubsection{Architecture}

\begin{figure}[h!]
\centering
\includegraphics[width=\columnwidth]{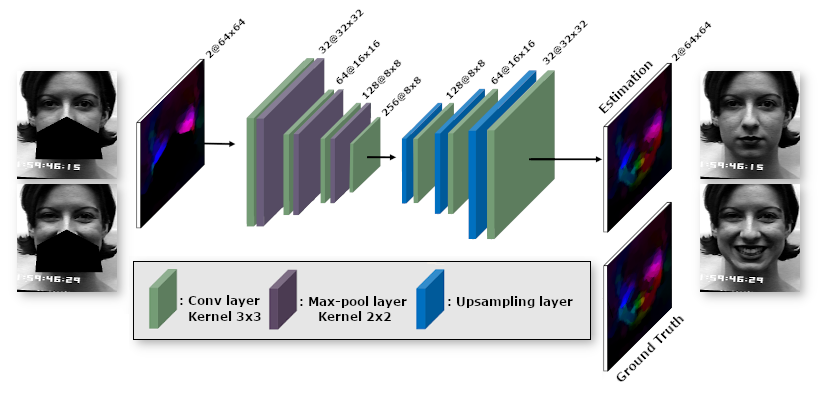}
\caption{Overview of our approach: a symmetric auto-encoder architecture processes the optical flow calculated from images of occluded faces and outputs the reconstructed optical flow. The auto-encoder is trained by computing a loss based on the comparison between the generated flow and the ground truth flow. The ground truth is known by calculating the optical flow between the same images as for the input but without occlusion. } 
\label{fig:auto-encoder}
\end{figure}

The architecture of the auto-encoder used in our approach is composed of successive convolutional and max-pooling layers for the encoder part and convolutional and up-sampling layers for the decoder part as illustrated in Fig. \ref{fig:auto-encoder}. 

The proposed architecture is symmetrical, which implies that the encoder part is inversely identical to the decoder part. Each convolution layer of a 3x3 kernel is followed by a ReLU and a two factor max pooling. The last convolution of the decoder part produces the same feature channels of input optical flow in order to get back to the initial dimensions. After the last convolution, there is no use of ReLU because the goal is to reproduce an optical flow which can be composed of negative components.

In general, after each up-sampling operation, the tensor goes through two successive layers of convolution to increase the intensity. Compared to classical architectures, only one convolution layer is used in our architecture. If we try to build very faithfully the missing optical flow, there is more chance to include motion discontinuities within the estimated region that may have a negative impact on the facial expression analysis. Whereas using a single layer, the model reconstructs a coarser optical flow that tends to rely heavily on a continuity of facial movement observed in the unoccluded regions, and generates less detail to facilitate facial expression analysis.

\subsubsection{Proposed loss functions}

Different loss functions have been used in order to find the most appropriate one in order to reconstruct the optical flow, while preserving the information of the facial expression in order to optimize the facial expression recognition step.

\textit{MSE loss:}
The mean squared error (MSE) is a frequently used loss with a denoising auto-encoder. The MSE loss is calculated by comparing each pixel of the output of the network with the pixel with the same coordinates in the ground truth image.

\textit{Wing loss:}
The wing loss was first proposed for the task of landmarks localisation \cite{feng2018wing}. This loss has the specificity of paying more attention to medium and small errors, which allows to have a more accurate prediction than with the MSE loss. We assume that this loss should generate a more accurate and detailed reconstruction of the optical flow since it corrects smaller errors.

\textit{Endpoint loss:}
Finally, we use the endpoint loss as it is a standard error measure for the evaluation of optical flows \cite{dosovitskiy2015}. Endpoint error is usually used to measure the error of calculation of an optical flow compared to the ground truth.

\subsection{Facial Expression Classification}
Facial expression recognition approaches are nowadays mostly based on deep learning approaches, and they have proven their effectiveness and give state-of-the-art results. Among the different architectures that are proposed in the literature, CNN architectures seem to be the mostly used architectures for static facial expression recognition \cite{li2020}. In order to add time, Allaert et al. \cite{allaert2019} proposed to feed a CNN architecture with optical flows. We propose in this paper to build the same architecture as in \cite{allaert2019} as it has been already developed in the context of optical flow classification for facial expression recognition. This architecture is not necessarily the state-of-the-art architecture but, in this paper, we do not focus on the performance of the recognition classifier but on the ability of our method to compensate under the effect of occlusions on the performance of the classifier.

\section{Experimental Evaluations}
\label{evaluation}
For our experimental evaluations, we propose, first, to analyse the impact of the size of the input image to correctly recognize facial expressions using the proposed CNN architecture. Once the optimal size has been determined, we then propose to evaluate the entire method by analysing the impact of several other meta-parameters. First, we evaluate the impact of the loss function used to train the auto-encoder. Second, we explore several strategies to calculate the optical flows. Third, we explore data augmentation techniques by combining different datasets for the training step of the auto-encoder and we study the impact on the reconstruction of the optical flow. Finally, when the different parameters and training data are fixed, we compare our results with the ones produced by the state-of-the-art methods.

\subsection{Experimental protocol}
In this section, we define the details of implementation for the evaluation protocol of the facial expression recognition and the reconstruction. In this section, we first expose the dataset selected for the evaluation before explaining, in one hand, the experimental protocol proposed to evaluate the CNN architecture and, on the other hand, the entire process including the reconstruction step. For this goal, the reconstructed optical flows are evaluated according to the recognition rate they can obtain with the CNN architecture.

\subsubsection{Dataset}
The CK+ dataset is used to evaluate our approach. The CK+ dataset is composed of 374 video sequences labelled with 6 facial expressions. The CK+ dataset has the advantage of being completely controlled in order to focus on the challenge of occlusion. Moreover this dataset is majoritarily used in the literature \cite{dapogny2018confidence,poux2020facial,li2018,cornejo2018emotion} for exploring the challenge of occlusion and allows a comparison with other methods. 

\subsubsection{Facial Expression Recognition}
The facial expression recognition is evaluated based on a CNN architecture trained to recognize facial expressions. The CNN is composed of three successive layers of convolution, ReLu activation and max pooling and the architecture ends with connected layers for the classification. The loss function used for training this CNN is the cross entropy loss. This architecture is the one proposed by Allaert et al. \cite{allaert2019} which has the advantage of being evaluated directly with optical flows for the input. The meta-parameters used for this architecture are frequently used parameters. We are aware that this architecture with these meta-parameters are not necessarily the state-of-the-art ones but our goal is not to improve CNN architectures to get the best recognition rates but to compensate the corruption of occlusions by reconstructing data to replace in a controlled environment.

\subsubsection{Evaluation protocol}
\begin{figure}[h!]
    \centering
    \includegraphics[width=3.5in]{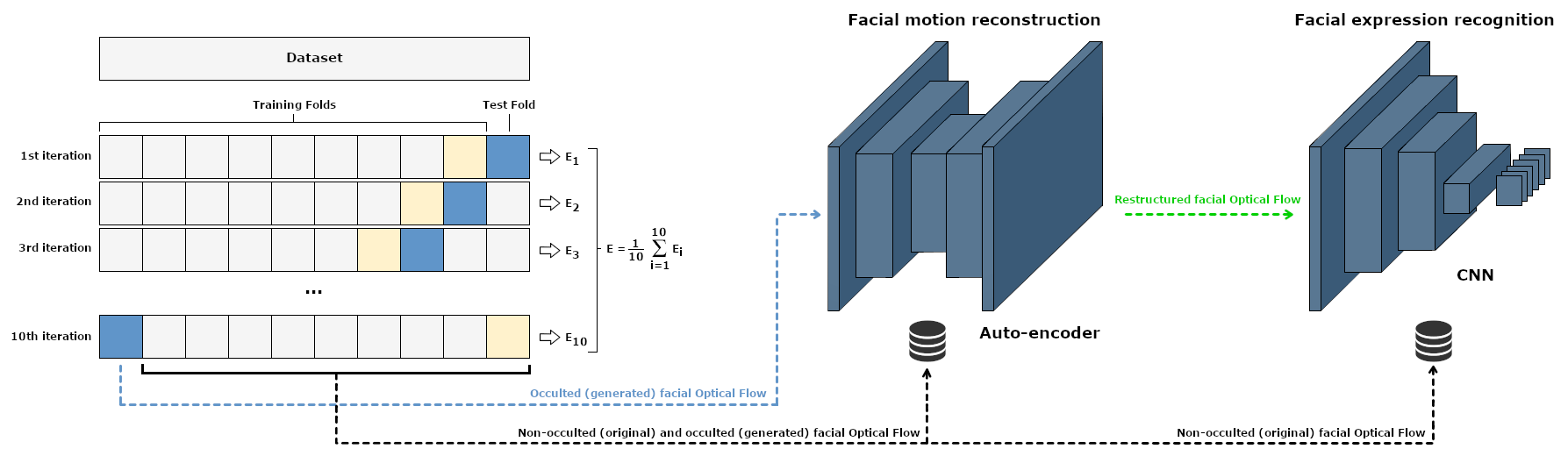}
    \caption{Ten folds cross protocol used to evaluate our method. For each experimentation, eight folds are used for training, one fold for validation and the last one for testing. The final accuracy is the mean over the number of 10-fold iterations. We also use the same partitions for both reconstruction and recognition.}
    \label{fig:folds}
\end{figure}

In order to evaluate the recognition step, we choose ten folds cross validation protocol which is mostly used in the literature \cite{dapogny2018confidence,poux2020facial,li2018}. The dataset is cut in ten stratified folds which contains proportional number of each facial expression compared with the original dataset. As illustrated in Fig. \ref{fig:folds}, the ten folds are successively used as test fold. For each experimentation, eight folds are used to train the network, one fold is used for validation in order to select the model and the last fold is used for testing. This cross validation protocol is first used to evaluate the facial expression recognition part in order to calculate the baseline without occlusion and to save the models used for the next experimentations. In a second part, the same protocol with the same folds are used to train the auto-encoder used for the reconstruction step. Eight folds are used to train the auto-encoder, one fold is used to validate the model and the last fold is used to test. For this last test, the images from the testing fold are reconstructed by the auto-encoder and the recognition rate of the reconstructed fold is calculated according to the corresponding CNN model previously trained. 

\textit{Experimental protocol for the reconstruction step}
For the evaluation of the reconstruction step, an occlusion $O$ is applied on all images of CK+. As illustrated in Fig. \ref{fig:folds}, we train the network with eight folds and validate on the last one. We ensure with this protocol to learn the auto-encoder and the CNN architectures on the same data for each experimentation. The test fold is also the same to ensure that the evaluation of the reconstructed data has not been already seen in the training step of the CNN model. We evaluate the entire process once for each fold to average the results obtained on the ten folds. After each epoch, the network is evaluated on the validation fold and the result model is selected according to the best results on the validation fold. The result model is then applied on the test fold.

\textit{Experimental protocol for expression recognition}
For the evaluation, The split in fold used to evaluate the expression recognition is used with the same process : eight folds are used to train the CNN, one fold is used for validation and the model is selected according to the best recognition rate on the validation fold. The last fold is used for testing. Successively, the folds are changed to test and validate on the ten different folds. The reported accuracy is the mean accuracy of the ten testing folds.

\subsection{Expression recognition without reconstruction}
The recognition step is first evaluated in order to find an appropriate compromise between the size of the input and the results of the recognition network. Indeed, smaller the image is, smaller is the number of meta-parameters for the auto-encoder for the reconstruction step. Thus, we are searching for the minimal size of input image required to keep optimal recognition rates. Besides, this first evaluation allows to give baselines in order to get the results without occlusion in one hand, and the results obtained with occlusions without using the proposed reconstruction method.

In this section, we evaluate the results obtained for several sizes of optical flows. We have evaluated the recognition performances on a wide range of input sizes : 24x24, 48x48, 64x64, 96x96 and 128x128. In order to have accurate results with a reduced impact of random initialisation, we evaluate each size on 100 different seeds. The results are obtained by calculating the median and the average results of the 100 scores. Once the optimal size is found, we fix the seed in order to train a CNN which allows a recognition rate representative with average results.

\subsubsection{Evaluation of expression recognition rate according to different sizes}
\begin{figure}[h!]
\centering
\includegraphics[width=0.4\columnwidth]{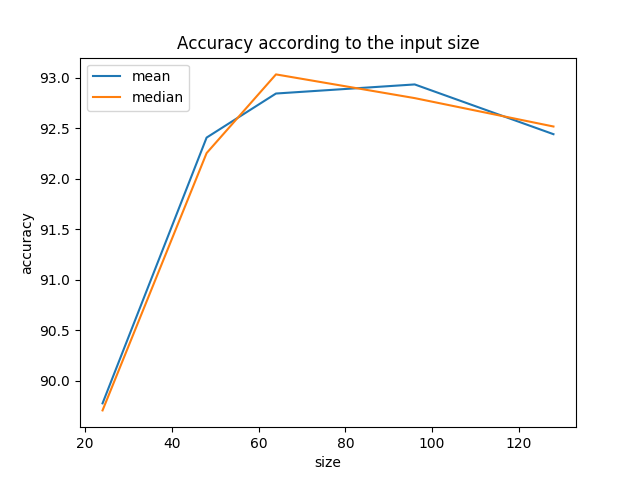}
\includegraphics[width=0.4\columnwidth]{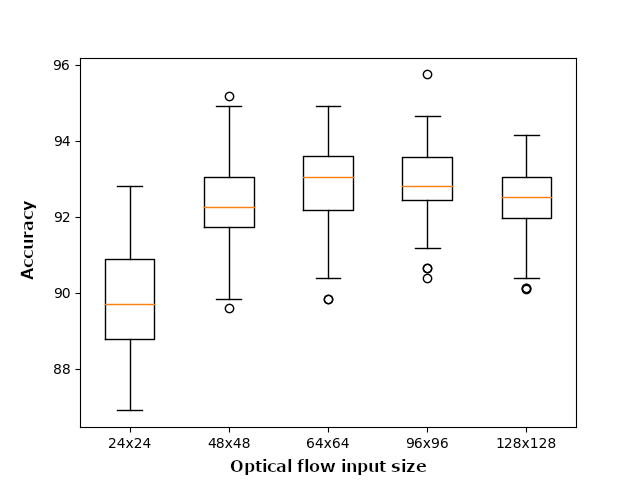}
\caption{Recognition rate of the CNN architecture in terms of the input size (after reduction of the size of the optical flow). This evaluation is conducted on 100 different seeds. The graph on the left shows the average and the median of those 100 experiments. The graph on the right shows the box plots in terms of the sizes of the optical flow. The flier points are the results out of the whiskers.}
\label{sizes}
\end{figure}

The recognition architecture is evaluated with different input sizes for the optical flow according to the above mentioned protocol. The results obtained for this evaluation are presented in Fig. \ref{sizes}.

As seen in this figure, the normalized optical flow give enough information for the architecture to correctly recognize the expressions with a small size. The best results are obtained between sizes of 64x64 and 96x96. According to the results presented in this graph, we can consider that a higher size does not seem to give more information but the input optical flow is more complex and more training data is needed, which can explain the fact that the results are decreasing. 

According to these results, 64x64 seems to be a reasonable size. The seed is then fixed for all the following experimentations by choosing a seed which allows an average result for the recognition rate. Once the seed is fixed, the CNN is trained and this trained model is used to evaluate the reconstruction step according to the different studied parameters. Based on these criteria, the baselines with and without occlusions can be calculated. These baselines are presented in Table \ref{baselineScores}. To calculate these scores, the CNN has been trained and validated on non-occluded faces and tested on one hand with non-occluded faces and, on the other hand, on occluded faces. These scores allow us to compare the results obtained with and without the proposed method to reconstruct the optical flows. 

\begin{table}[h]
\caption{Baseline calculated from non-occluded faces and different occlusions with optical flow of size 64x64.}
\label{baselineScores}
\centering
\begin{tabular}{|c|c|c|c|c|}
\hline
 \includegraphics[width=0.62in]{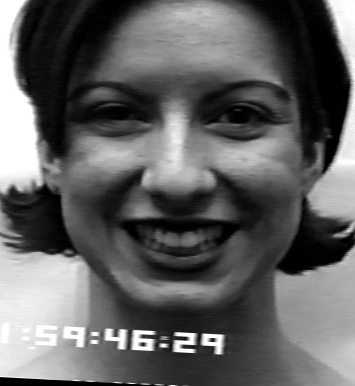} & \includegraphics[width=0.62in]{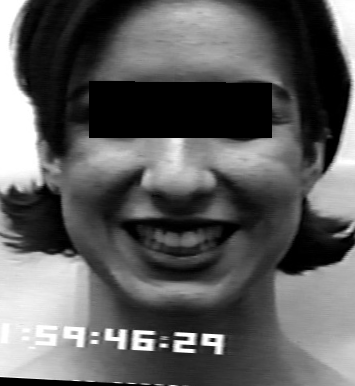} & \includegraphics[width=0.62in]{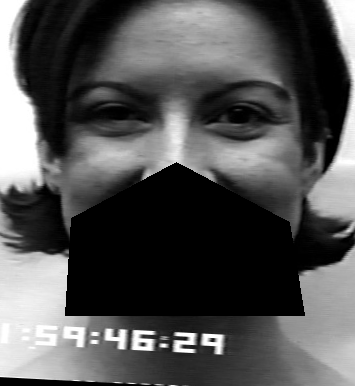} & \includegraphics[width=0.62in]{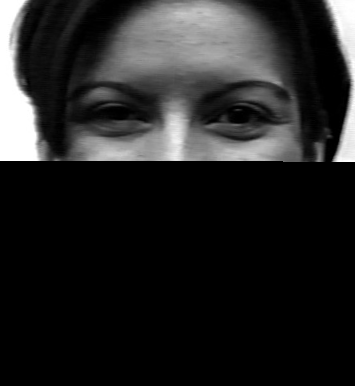} \\ \hline
92,8\% & 73,8\% & 71,1\% & 46,8\% \\ \hline
\end{tabular}
\end{table}

\subsection{Expression recognition with reconstruction}
 In the first part of the evaluation, we have fixed the input optical flow image size and the seed used for the expression recognition step. This step gives us a baseline which indicates the results we can obtain without occlusion and the results obtained with a training on non-occluded faces and testing on occluded faces. In this section, we evaluate the reconstruction step by comparing the results obtained with reconstructed optical flows with the above-mentioned results. In order to do that, we first have to find optimal parameters for the entire process. After explaining the experimental protocol for these evaluations, we evaluate the different above-mentioned parameters. When all the optimal parameters are fixed, the results are compared with state-of-the-art results.

\subsubsection{Evaluation of the Proposed method w.r.t. to different loss functions}
Different loss functions have been proposed to train the auto-encoder to reconstruct the corrupted optical flows : 
\begin{itemize}
    \item the MSE loss which is a mostly used loss function,
    \item the wing loss which pays more attention to small errors and should give more accurate results,
    \item the endpoint loss which is a standard error measure for the evaluation of optical flows.
\end{itemize}

These three loss functions are compared in this section to find the most appropriate for the proposed auto-encoder. The above-mentionned experimental protocol is applied with the CK+ dataset and we consider optical flows calculated between the first and the last frame of each sequence. Tables \ref{accLoss} and \ref{diffLoss} present the results by respectively presenting the accuracy of expression recognition on reconstructed optical flows and the average gain for the several occlusions compared to the results without reconstruction. As observed on those tables, on average, the endpoint seem to be more adapted to this context. For the following experiments, we, thus, fix the endpoint loss and give only results with this loss.

\begin{table}[h]
\caption{Comparison of the accuracy for expression recognition obtained with reconstructed optical flow according to different loss functions for training the reconstruction auto-encoder.}
\label{accLoss}
\centering
\begin{tabular}{|c|c|c|c|}
\cline{2-4}
\multicolumn{1}{c|}{} & Eyes occ. & Mouth occ. & Lower part occ. \\ \hline
MSE & 86.2\% & 74.0\% & 67.3\% \\ \hline
Wing & 86.1\% & 79.5\% & 69.0\% \\ \hline
EndPoint & \textbf{87.1\%} & \textbf{80.1\%} & \textbf{70.2\%} \\ \hline
\end{tabular}
\end{table}

\begin{table}[h]
\fontsize{7}{8}\selectfont
\caption{Comparison of the loss/gain obtained for expression recognition obtained with reconstructed optical flow according to different loss functions for training the reconstruction auto-encoder.}
\label{diffLoss}
\centering
\begin{tabular}{|c|c|c|c|c|}
\cline{2-5}
\multicolumn{1}{c|}{} & Eyes occ. & Mouth occ. & Lower part occ. & Average gain \\ \hline
MSE & +12.4 & +2.9 & +20.5 & +11.9 \\ \hline
Wing & +12.3 & +8.4 & +22.2 & +14.3 \\ \hline
EndPoint & +13.3 & +9.0 & +23.4 & +15.2 \\ \hline
\end{tabular}
\end{table}

\subsubsection{Evaluation of the Proposed method w.r.t. skip connections}
\label{skip}

\begin{figure}[h!]
    \centering
    \includegraphics[width=0.5\linewidth]{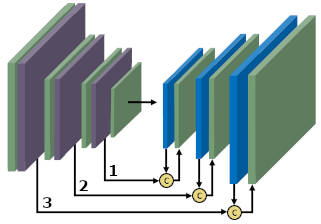}
    \caption{Illustration of the different skip connections studied.}
    \label{fig:skip_connections}
\end{figure}

We propose to add skip connections to the network to merge the information from the previous tensor and the oversampling tensor, which allows the details of the previous layers to be used completely. Indeed, some available information in the visible regions of the face may be lost in the bottleneck of the network. For this goal, we propose to add skip connections to reuse some features from the encoder. In order to analyze the impact of these connection, we propose to study the different possible connection combinations to highlight the importance of these connections, illustrated in Fig. \ref{fig:skip_connections}, and the importance of their role in the reconstruction of the optical flow for the recognition of facial expressions.

Inspired by the U-net architecture \cite{ronneberger2015u}, rather than using element addition to merge the information from the previous tensor and the oversampling tensor, our architecture concatenates them respecting the channel dimension in the decoder part. This recalls the details of the previous layers in the reconstruction process.

Table \ref{tab:skip_connections} shows that skip connections have a positive impact on the final results. These results tend to show that skip connections allow a more detailed reconstruction of the optical flows. This table also shows that skip connections on all layers seems to be the most appropriate. For the following experiments, the architecture is fixed by adding skip connections on the three layers.

\begin{table}[h!]
    \centering
    \begin{tabular}{|c|c|c|c|}
    \cline{2-4}
    \multicolumn{1}{c|}{} &  Eyes occ. & Mouth occ. & Lower part occ. \\
    \hline
     / & 87.1\% & 80.1\% & 70.2\%\\
    \hline 
    1 & 88.4\% & 81.1\% & 71.1\%\\
    \hline
    2 & 89.3\% & 81.8\% & 71.9\%\\
    \hline
    3 & 89.4\% & 81.0\% & 70.8\%\\
    \hline
    1+2 & 88.9\% & 82.6\% & 71.4\% \\
    \hline
    1+3 & 89.3\% & 82.5\% & 71.5\% \\
    \hline
    2+3 & 89.5\% & 82.6\% & 72.1\% \\
    \hline
    1+2+3 & \textbf{89.6\%} & \textbf{83.1\%} & \textbf{72.5\%} \\
    \hline
    \end{tabular}
    \caption{Impact of skip connections in the auto-encoder performances. The results correspond to the final facial expression recognition rate after using the entire process. The different skip connections correspond to the numbers illustrated in Fig. \ref{fig:skip_connections}}
    \label{tab:skip_connections}
\end{table}

\subsubsection{Evaluation of the Proposed method w.r.t. the optical flow computation strategy}
\begin{figure}[h!]
\includegraphics[width=3.5in]{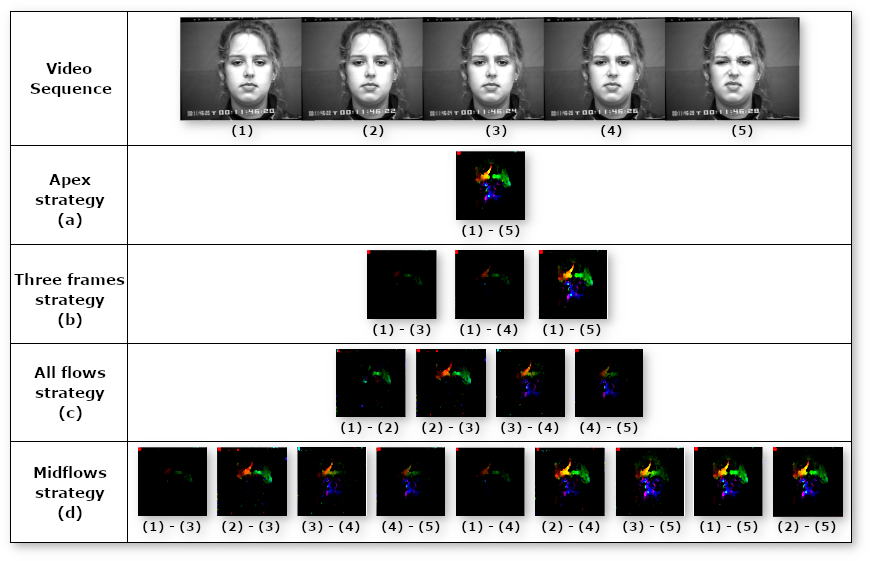}
\centering
\caption{Five ways of calculating the optical flows are proposed. The first line illustrates a part of a video sequence from the CK+ dataset. Each frame has been numbered from 1 to 5 to facilitate the comprehension. The next lines of the figure illustrate the different strategies. Under each visualisation of optical flow, the number of the frames used to compute the flow written as ($prvs$)-($next$). (a) The $apex$ strategy only calculates the optical flow between the first and the last frame of the sequence. (b) The three frames strategy considers the first frame and the three last frames to compute different optical flows. This strategy aims to add flows by considering the last three frames as three apex frames. (c) The all flows strategy computes all consecutive flows. The flows and apex strategy which combines the optical flows calculated with the strategies (a) and (c). (d) The midflows strategy calculates all possible flows where each next frame of the couple (prvs, next) is chosen from the second half part of the sequence.}
\label{optFlow}
\end{figure}

The input data of the auto-encoder is given by optical flows calculated between two images. We have selected the Farnebäck method to compute optical flows as Allaert et al. \cite{allaert2019} show that this method is particularly adapted for facial expressions.
In order to ensure sufficient varieties of optical flow intensities on different expressions, we propose to study and compare three ways of calculating the optical flows for each video sequence illustrated in Fig. \ref{optFlow}.
Let $prvs$ and $next$ two frames used to calculate the optical flow and $n$ the number of frames in a sequence $S$.
\begin{itemize}
\item Apex flow: The first proposition, represented on line (a) of the Fig. \ref{optFlow}, calculates only one optical flow for each sequence $S$ with $prvs$ the first frame of the sequence and $next$ the last frame.
\item All flows : The second solution to calculate the optical flows of a sequence $S$, represented on line (b) of Fig. \ref{optFlow}, calculates all the ($prvs$,$next$) where $prvs$ is a $t^{th}$ frame and $next$ the $(t+1)^{th}$ for $t \in [1,n]$.
\item Flows and apex: The third proposition is a combination of the two precedent solutions. This proposition allows a training containing flows with small and important intensities.
\item Mid flows: The last proposition, illustrated on line (c) of Fig.\ref{optFlow}, calculates all the possible optical flows between each $prvs$ and $next$ with $prvs$ the $t^{th}$ where $t\in[1,n-1]$ and $next \in [max(t+1,n/2),n]$
\end{itemize}

Table \ref{loaders} gives the results obtained with the endpoint loss according to the several strategies. As we can see in this table, although $apex$ generates few optical flows, it is already a good strategy to train the auto-encoder. We can also notice that $all flows$ strategy is not adapted, which can be totally explained by the fact that this strategy only generates optical flow with small movements as it considers consecutive frames. However, the facial expression recognition step considers the entire movement calculated between the first and the last frame of the sequence, which is a much bigger movement. $Flows and apex$ gives higher scores than $all flows$ as it also considers big movements between first and last frame. Nevertheless, all the small movements added with this strategy only reduce the ability of the auto-encoder to reconstruct big movements. Finally, $mid flows$ strategy seem to be the most appropriate with the highest scores for each occlusion. This last observation is easily understandable as this strategy generates the most important amount of optical flows for training. Moreover, this strategy makes sure that the intensity of the movement is not too small by avoiding the first middle part of the video sequence.
	
\begin{table}[h]
\caption{Impact of the several strategies according to the training input for the studied occlusions.}
\label{loaders}
\centering
\begin{tabular}{|l|c|c|c|c|}
\cline{2-4}
\multicolumn{1}{c|}{} & Eyes occ. & Mouth occ. & Lower part occ.  \\ \hline
Apex flow & 89.6\% & 83.1\% & 72.5\% \\ \hline
Three frames & 90.2\% & 83.6\% & 74.4\% \\ \hline
All flows & 85.5\% & 77.6\% & 66.8\% \\ \hline
Flows and apex & 89.6\% & 79.1\% & 71.8\% \\ \hline
Mid flows & \textbf{91.1\%} & \textbf{85.9\%} & \textbf{75.2\%} \\ \hline
\end{tabular}
\end{table}	
		
\subsubsection{Performances comparison with the literature}
We compare our approach to the state-of-the-art methods, and attempt to explain why movement reconstruction improves expression recognition in the presence of occlusions. In the evaluation section (Sec. 4), the different proposed parameters were fixed by taking the optimal ones. Here, the endpoint loss is used and the $mid flows$ strategy is kept. With these parameters fixed, we can therefore compare the results of the proposed approach with other methods.

Table \ref{soa_comparison} compares the results obtained using our reconstruction step with other state-of-the-art methods which has been evaluated on the CK+ dataset. In this table, the best performance is highlighted in bold and the second best is underlined. In order to quantify the impact of occlusion on the performance of the different approaches, the gap between the baseline `without occlusion' and in the presence of different types of occlusions is reported. To easily compare the different approaches, the cumulative gap obtained from the `eyes occlusion' and the `mouth occlusion' is computed. One can notice from the table that the behavior is the same for all methods : occlusion of the eyes has less impact on the performance than the occlusion of the mouth regions. We can especially notice that the results achieved with our method are very competitive. Indeed, our approach achieves the second best performance for the eyes occlusion and the best results for mouth and `lower part' occlusions which have the greatest impact on facial expression recognition. The results obtained with our approach with the lower part occlusion (a more important area of the face) are competitive with occlusions of the mouth of the state-of-the-art methods.

\begin{table}[h]
\caption{Comparison of the results of our proposed method and the state-of-the-art results on the CK+ dataset for eyes, mouth and lower part occlusion. This table illustrates the impact of different types of occlusions on the performance of different techniques, and shows the gap in performance between `no occlusion' and different types of occlusion. A cumulative gap of `eyes and mouth' occlusions is also shown in the fourth row of the table to easily compare the performance of the different approaches.}
\label{soa_comparison}
\centering
\fontsize{6}{8}\selectfont
\begin{tabular}{|l|c|c|c|c|c|}
\cline{2-6}
\multicolumn{1}{c|}{} & \includegraphics[width=0.05\textwidth]{img/newCropNxtnoOcclusion_black.png} & \includegraphics[width=0.05\textwidth]{img/newCropNxteyes1_black.png} & \includegraphics[width=0.05\textwidth]{img/newCropNxtmouth2_black.png} & & \includegraphics[width=0.05\textwidth]{img/newCropNxtmouth3_black.png} \\
\hline
Methods & No occ. & Eyes occ. & Mouth occ. & Gap & Lower part occ. \\ \hline
Huang et al.  & 93.2\% & \bf{93\%} & 73.5\% & & \\ \cite{huang2012towards} & &  \bf{-0.2\%} & -19.7\% & -19.9\% & - \\ \hline
Dapogny et al. & \bf{93.4\%} & 76\% & 67.1\% & & \\ \cite{dapogny2018confidence} &  & -17.4\% & - 26.3\% & -43.7\% & - \\ \hline
Poux et al. & 91.3\% & 88.8\% & \underline{79\%} & & \underline{73.4\%} \\ \cite{poux2020facial} & & -2.5\% &  \underline{-12.3\%} & -14.8\% & \underline{-17.9\%} \\ \hline
\hline
Our method & 92.8\% & \underline{91.1\%} & \bf{85.9\%} & & \bf{75.2\%} \\
& & \underline{-1.7\%} & \bf{-6.9\%} & \bf{-8.6\%} & \bf{-17.6\%} \\ \hline
\end{tabular}
\end{table}

The difference in performance may be explained by the features on which these different approaches are based (e.g., static or dynamic). Indeed, in addition to the use of more complex occlusions based on salt and pepper boxes, Dapogny et al. \cite{dapogny2018confidence} proposed an approach based on static images which may explain the gap obtained with occlusions because of the loss of the temporal information. Huang et al. \cite{huang2012towards} approach uses texture and shape dynamic features which can explain the fact that the solution is more robust to occlusions thanks to the temporal information. Nevertheless, the features proposed by Huang et al. are not based on dense movements and the remaining information on the cheeks, when the mouth is occluded, may not be sufficient as the cheeks are not textured. Moreover, in Poux et al. \cite{poux2020facial} a solution based on dense movements using an optical flow descriptor was proposed, which allows to exploit the property of propagation of movements and can explain the gain obtained, especially under mouth occlusion. The important difference of the proposed approach with the one proposed in Poux et al. (which also holds for the case of the two other methods) is that, these three solutions rely on a sub-region approach which focuses its attention on the visible regions, and gives less importance to the occluded parts. The difference between the results of our approach and especially the one of Poux et al. \cite{poux2020facial} (which is also based on optical flow) may be explained by the fact that our approach takes advantage of all visible parts for reconstruction, and does not only rely on some of the visible regions of the face. Moreover, our solution, which is based on the movement (by reconstructing optical flows), exploits the property of similarity of the movement between different persons.

These results show the importance of the reconstruction of the optical flow to handle occlusions for facial expression recognition. We can also notice that the results of our approach, especially for eyes occlusion, may be improved. In this paper, we proposed a basic architecture based on an auto-encoder, which can easily be improved with a more complex architecture. Moreover, this approach relies on a reconstruction trained with the loss between the reconstructed optical flow and the ground truth optical flow, calculated on images without occlusion, while the goal is to classify facial expression under occlusion. The solution can be improved by modifying the loss in order to ensure that the information about the facial expression is preserved. In this case, a less qualitative reconstruction may be more informative if the information about the expression is well reconstructed.

There is a caveat to the above comparisons and justification in terms of training and testing protocols that are adopted by the different approaches, and which do not lead to a totally fair and impartial comparison. First, Huang et al. and Dapogny et al. evaluated their approaches on 8 facial expressions, including neutral and contempt, while we are only using 6 facial expressions. Second there is a difference between approaches in the split of the datasets into training and testing datasets. Indeed, Dapogny et al. evaluated their approach with an Out of Bag protocol, Huang et al. used a Leave One Subject Out method and we use a 10-folds cross validation.  We believe that establishing a clear protocol for occlusion generation and evaluation, as presented in this work, will support finer and fairer comparisons in the future.

\section{Conclusion}
\label{subsec:conclusion}
In this paper, we propose a novel approach to recognize facial expressions in presence of partial facial occlusions. The proposed approach consists in reconstructing directly calculated optical flows corrupted by a partial occlusion before using these reconstructed optical flows for the recognition step. In order to do that, we propose a denoising auto-encoder trained to reconstruct corrupted optical flow by using a loss which compares the reconstruction with the ground truth optical flow calculated on images without occlusions. To find the best way to reconstruct optical flows, we proposed to compare different loss functions and different ways to calculate the training optical flows. The several proposed studies have shown that the endpoint loss and the $mid flows$ strategy which considers a wide range of intensity of optical flows seem to be the more suitable choices for this particular task. When comparing the obtained results with state-of-the-art results, we can see that the proposed method is really competitive. In a future work, we want to study the impact of dynamic occlusions on this method in order to propose a solution which takes into account these kind of more natural occlusions.
To go further, our future work will be to study occlusions caused by head pose variations. In this case, in addition to the loss of movement behind the occlusion there is a noisy movement caused by the head.



\ifCLASSOPTIONcaptionsoff
  \newpage
\fi

\bibliographystyle{plain}
\bibliography{biblio}

\begin{IEEEbiography}
[{\includegraphics[width=1in,height=1.25in,clip,keepaspectratio]{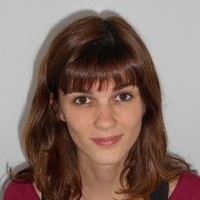}}]
{Delphine Poux}
received his MS degree on Complex Model, Algorithms and Data in Computer Science from
the University of Lille, France. She is currently a Ph.D. student at the Computer Science Laboratory in Lille
(CRIStAL). Her research interests include computer vision and affective computing.
\end{IEEEbiography}

\begin{IEEEbiography}
[{\includegraphics[width=1in,height=1.25in,clip,keepaspectratio]{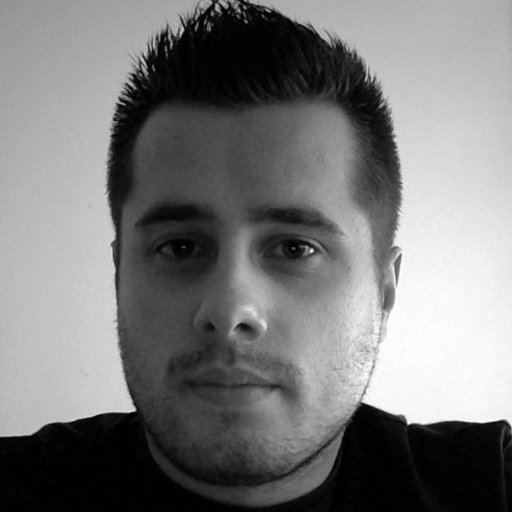}}]
{Benjamin Allaert}
received his MS degree on Image,Vision and Interaction and his Ph.D. on analysis of
facial expressions in video flows in Computer Science from the University of Lille, France. He is currently
a research engineer at the Computer Science Laboratory in Lille (CRIStAL). His research interests include
computer vision and affective computing, and current focus of interest is the automatic analysis of human
behavior.
\end{IEEEbiography}


\begin{IEEEbiography}
[{\includegraphics[width=1in,height=1.25in,clip,keepaspectratio]{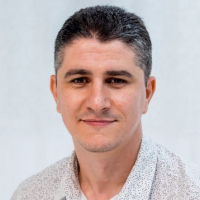}}]
{Nacim Ihaddadene}
e is an Assistant Professor at the engineer school Isen, France. He received his MS degree
on software architectures and his Ph.D. on Extracting business process models from event logs in Computer
Science from respectively the University of Nantes and the University of Lille. Then, he integrated the HEI
ISA ISEN Group. Since, he extended his research to human behavior understanding.
\end{IEEEbiography}

\begin{IEEEbiography}
[{\includegraphics[width=1in,height=1.25in,clip,keepaspectratio]{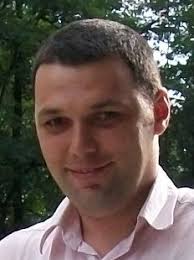}}]
{Ioan Marius Bilasco}
is an Assistant Professor at the University of Lille, France, since 2009. He received his
MS degree on multimedia adaptation and his Ph.D. on semantic adaptation of 3D data in Computer Science
from the University Joseph Fourier in Grenoble. In 2008, he integrated the Computer Science Laboratory in
Lille (CRIStAL, formerly LIFL) as an expert in metadata modeling activities. Since, he extended his research
to facial expressions and human behavior analysis.
\end{IEEEbiography}

\begin{IEEEbiography}
[{\includegraphics[width=1in,height=1.25in,clip,keepaspectratio]{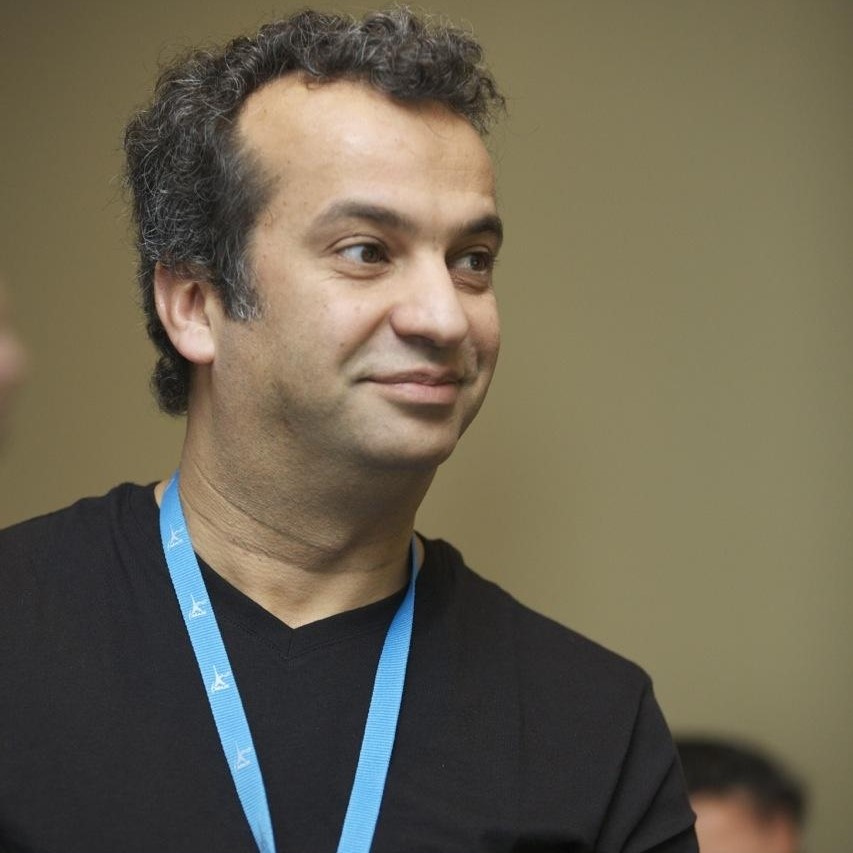}}]
{Chaabane Djeraba}
obtained a MS and Ph.D. degrees in Computer Science, from respectively the “Pierre
Mendes France” University of Grenoble (France) and the “Claude Bernard” University of Lyon (France). He
then became an Assistant and Associate Professor in Computer Science at the Polytechnic School of Nantes
University, France. Since 2003, he has been a full Professor at the University of Lille. His current research
interests cover the extraction of human behavior related information from videos, as well as multimedia
indexing and mining.
\end{IEEEbiography}

\begin{IEEEbiography}
[{\includegraphics[width=1in,height=1.25in,clip,keepaspectratio]{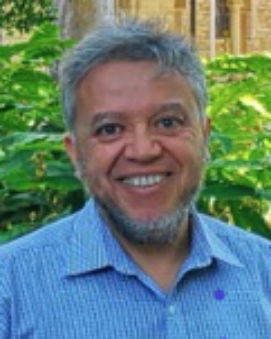}}]
{Mohammed Bennamoun}
is Winthrop professor with the Department of Computer Science and Software Engineering at UWA and, a researcher in computer vision, machine/deep learning, robotics, and signal/speech processing. He has published four books (available on Amazon), one edited book, one Encyclopedia article, 14 book chapters, more than 120 journal papers, more than 250 conference publications, 16 invited \& keynote publications. His h-index is 53 and his number of citations is 12,000+ (Google Scholar). He was awarded 65+ competitive research grants, from the Australian Research Council, and numerous other Government, UWA and industry Research Grants. He successfully supervised 26+ PhD students to completion. He won the Best Supervisor of the Year Award at QUT (1998), and received award for research supervision at UWA (2008 \& 2016) and Vice-Chancellor Award for mentorship (2016). He delivered conference tutorials at major conferences, including: IEEE Computer Vision and Pattern Recognition (CVPR 2016), Interspeech 2014, IEEE International Conference on Acoustics Speech and Signal Processing (ICASSP) and European Conference on Computer Vision (ECCV). He was also invited to give a Tutorial at an International Summer School on Deep Learning (DeepLearn 2017).
\end{IEEEbiography}




\end{document}